\documentclass[11pt]{article}
\usepackage[a4paper, total={7in, 10in}]{geometry}

\usepackage{graphicx}
\usepackage{helvet}
\usepackage{authblk}
\usepackage{xcolor}
\usepackage[colorlinks=true,linkcolor=blue,citecolor=blue,urlcolor=blue,allbordercolors=white]{hyperref} 
\usepackage{amsmath} 
\usepackage{amssymb} 
\usepackage[super,comma,sort&compress]{natbib}
\makeatletter
\renewcommand{\maketitle}{\bgroup\setlength{\parindent}{0pt}
\begin{flushleft}
  \textbf{\@title}
  
  \@author
\end{flushleft}\egroup}
\makeatother

\title{\Large Reinforcement learning in densely recurrent biological networks}
\date{}

\author[1,2,*]{Miles Walter Churchland}
\author[1,**]{Jordi Garcia-Ojalvo}

\affil[1]{Department of Medicine and Life Sciences, Universitat Pompeu Fabra, Barcelona Biomedical Research Park (PRBB), Dr Aiguader, 88, 08003 Barcelona, Spain}
\affil[2]{Mathematics and Computer Science Program, Reed College, 3203 SE Woodstock Boulevard, Portland, OR 97202, USA}

\affil[*]{Correspondence: mileschurchland@gmail.com}
\affil[**]{Correspondence: jordi.g.ojalvo@upf.edu}

\begin{document}

\maketitle

\section*{Abstract}

Training highly recurrent networks in continuous action spaces is a technical challenge: gradient-based methods suffer from exploding or vanishing gradients, while purely evolutionary searches converge slowly in high-dimensional weight spaces. 
We introduce a hybrid, derivative-free optimization framework that implements reinforcement learning by coupling global evolutionary exploration with local direct search exploitation. 
The method, termed ENOMAD (Evolutionary Nonlinear Optimization with Mesh Adaptive Direct search), is benchmarked on a suite of food-foraging tasks instantiated in the fully mapped neural connectome of the nematode \emph{Caenorhabditis elegans}.
Crucially, ENOMAD leverages biologically derived weight priors, letting it refine--rather than rebuild--the organism’s native circuitry.
Two algorithmic variants of the method are introduced, which lead to either small distributed adjustments of many weights, or larger changes on a limited number of weights.
Both variants significantly exceed the performance of the untrained connectome (in what can be interpreted as an example of transfer learning) and of existing training strategies.
These findings demonstrate that integrating evolutionary search with nonlinear optimization provides an efficient, biologically grounded strategy for specializing natural recurrent networks towards a specified set of tasks.

\section*{Keywords}

Biological networks, recurrent neural networks, evolutionary learning, reinforcement learning

\section{Introduction}
\label{sec:intro}

Biological networks are highly efficient information-processing systems \citep{tkacik2016,tkacik2025}.
Uncovering the principles behind these systems’ extraordinary functionality and adaptability requires identifying patterns in connection strengths (weights) and structural motifs that underpin evolved traits.
Biological networks are highly recurrent, however  \citep{gabalda2018,Casal2020}, which makes training the network weights a complicated undertaking.
On the positive side, living systems provide us with evolved priors, which offer biologically plausible initial weights and network structure.
Starting from these priors, task-specific optimization can fine-tune a real-life biological network for a chosen set of tasks, instead of training it from scratch.
Here we use the neural network of the nematode \emph{Caenorhabditis elegans} as a biological prior.
The connectome of \emph{C. elegans}, which contains 302 neurons and 3682 connections, is fully mapped, densely recurrent, and relatively small \citep{openworm}.
These characteristics make this organism an ideal testbed for assessing how effectively a training scheme can be at specializing a natural neural network for a specific task in a realistic environment.

An effective framework within which to implement the training strategy described above is reinforcement learning (RL), a machine-learning approach in which an agent improves by interacting with its environment to maximize a reward \citep{sutton1998reinforcement}.
Recent work has begun to use real connectomes as strong structural priors in RL strategies.
Ordinary Neural Circuits (ONCs) have been used, for instance, to re-parameterize the \emph{C.\ elegans}' wiring diagram and, using search-based RL, to control simulated and physical robots with two orders of magnitude fewer parameters than comparable deep nets \citep{Hasani2020}. 
At a larger scale, hierarchical and model-based RL have been proposed as computational accounts of insect navigation using the mushroom body neural circuit \citep{RLInsect2024}.
These examples highlight the promise —and challenge— of training highly recurrent natural networks.

Most canonical RL algorithms were designed for discrete action spaces and/or feed-forward networks (traditional recurrent neural networks fall into this category).  When the system (i) must rely on continuous motor commands and/or (ii) exhibits densely recurrent dynamics, many training approaches fail due to several reasons, including:
\begin{itemize}
    \item[RL1] \textbf{Unstable gradient estimates.}  Policy-gradient and actor-critic variants support continuous actions, but their gradients exhibit high variance. Thus, back-propagation through long recurrent paths suffers from vanishing or exploding gradients while also being computationally challenging \citep{Pascanu2013}.
    \item[RL2] \textbf{Action discretization bottleneck.}  Value-based methods such as Q-Learning \citep{watkins1992q}, deep Q-networks \citep{Mnih2015DQN}, and proximal policy optimization (PPO) algorithms \citep{Schulman2017PPO} scale exponentially with the number of discrete bins; hence continuous motor control becomes intractable.
\end{itemize}

A practical way around these problems is to abandon derivatives altogether and frame training as a black-box search in weight space.
Among the derivative-free methods, NOMAD (Non-linear Optimization with Mesh-Adaptive Direct search) is particularly appealing \citep{AuDe2006,nomad4paper}.
NOMAD is a mesh-search algorithm that, at every iteration, evaluates trial points in a shrinking mesh around the current best solution.
If any poll point improves the score, the mesh is refined (the step size is decreased) to focus the search; otherwise it is coarsened to promote exploration.
Because it is gradient-free, NOMAD naturally avoids issue \#RL1 above (unstable gradients).
As a black-box optimizer, NOMAD also sidesteps \#RL2’s action-discretization bottleneck by treating the task as a black-box function and tuning weights directly--no policy or value function needed.
Unlike reinforcement learning algorithms that aim to learn which actions yield higher rewards in various states, NOMAD treats the system as a black box.
It makes no assumptions about the structure of the problem--whether the action or state spaces are discrete, continuous, or highly irregular. Instead, it evaluates the fitness of each real-valued parameter vector over full trajectories, directly optimizing continuous controllers without ever discretizing the action space.

NOMAD is frequently combined with population-based optimizers such as Particle Swarm Optimization (PSO).
In several works, PSO has been combined with NOMAD either sequentially--PSO exploration followed by NOMAD refinement of the best solution\citep{HUMPHRIES2015242}--or in memetic hybrids where NOMAD is interwoven with PSO updates \citep{Lee2016}.
In spite of its advantages, the standard application of NOMAD method has several limitations:
\begin{itemize}
\item[N1] \textbf{Search space dimensionality.} NOMAD's convergence guarantees hold as long as a limited number of weights is optimized (typically less than 50 in standard implementations of the method).
However, we aim to optimize natural networks with thousands of parameters, two orders of magnitude larger than traditional NOMAD applications.
\item[N2] \textbf{Rugged, discontinuous reward spaces.} Traditional population-based optimization methods such as PSO assume smooth reward spaces. In densely recurrent networks, however, small synaptic tweaks can flip global dynamics, yielding a jagged reward surface.
This fact severely limits the usability of hybrid PSO+NOMAD methods for recurrent networks such as ours.
\end{itemize}

To address these two issues, we have developed a hybrid method that combines NOMAD with an evolutionary loop that tolerates discontinuities in reward space.
Additionally, we start training from a mapped natural network based on previous biological knowledge, rather than from random weights, and deliberately search in its vicinity. By assuming that there are strong local minima near our initialization point, we dramatically reduce our search space.

Evolutionary algorithms have a long history within the reinforcement learning field \citep{moriarty1999evolutionary}, and have also been applied to recurrent neural networks, where they have been used for instance to train gene regulatory networks within the reservoir computing framework \citep{gabalda2018} and infer the sign (excitatory/inhibitory) of all synaptic connections in the \emph{C. elegans} connectome \citep{Casal2020}.
By interweaving NOMAD refinement with every generation of an evolutionarily-driven global search, our method (i) scales to thousands of parameters, (ii) climbs steep, non-smooth fitness peaks, and (iii) respects strong biological priors, capabilities not exhibited by earlier hybrid schemes combining population-based optimization methods with NOMAD.
To demonstrate the effectiveness of our method, we trained the \emph{C.\ elegans} connectome across a suite of food-collection tasks, and compare it with existing approaches, including several evolutionary strategies (ES).

\section{Methods}
\label{sec:framework}

\subsection{Connectome}

The original connectome used in our study is sourced from the c302 project \citep{c302}, which is an initiative to model the neural network of the nematode \textit{C. elegans}.
The network is composed of several key components:
\begin{itemize}
    \item \textbf{Neurons}: The database includes all 302 identified neurons in the \textit{C. elegans} nervous system.
    Our model includes 300 neurons; CANL and CANR are omitted as they form only gap-junction contacts with excretory cells and have no chemical synapses with the rest of the connectome. The database also lists 96 muscle cells (only 68 of which are used to control the worm, as head muscle cells cause unexpected behavior due to our simple method of accumulating neuronal activity, presented in the next Sec.~\ref{sec:model}).
    \item \textbf{Synaptic connections}: The c302 connectome model contains 3682 synaptic connections, categorized into various types based on the nature of the synapse, including chemical synapses, gap junctions, and neuromuscular junctions (which link neurons to muscle cells).
    The connections are represented as directed edges in the network, indicating the direction of signal transmission.
    Gap junctions are considered bidirectional, with independent weights in the two directions for computational simplicity.
\item \textbf{Connection count (proxy weights)}: The dataset reports only the number of terminal contacts that neuron $i$ makes onto neuron $j$, not graded synaptic strengths. We therefore use that contact count $c_{ij}$ as an initial weight, $w_{ij}=c_{ij}$. Treating contact number as a proxy weight is standard in \emph{C.\ elegans} modeling \cite{c302project,Casal2020} and provides the biologically informed prior for our training. These weights are essential for accurate neural-dynamics simulation and form the starting point for ENOMAD optimization.

\end{itemize}

\subsection{Model}
\label{sec:model}

\textbf{Neuronal dynamics.} Neurons are assumed to follow a discrete-time representation of the integrate-and-fire dynamics with no leak:
\begin{equation}
V_k(t) = V_k(t-1)+\sum_{i \in \text{pre}_k} w_{ik} x_i(t-1),
\end{equation}
where $V_k(t)$ is the membrane potential of neuron $k$ at time $t$, $\text{pre}_k$ denotes the set of presynaptic neurons that innervate neuron $k$, $w_{ik}$ is the synaptic strength (weight) from neuron $i$ to neuron $k$,
and $x_{i}(t-1)$ is a binary representation of whether the presynaptic neuron $i$ fired at time $t-1$.
A neuron $i$ fires when its membrane potential $V_i$ reaches a threshold, after which $V_i$ is reset to 0.
All variables and parameters in the model use arbitrary units.
The model is simulated using parallelized trials \citep{moritz2018raydistributedframeworkemerging} along with Cython \citep{behnel2011cython}.
The code is publicly available in Github (\url{https://github.com/dsb-lab/C_Elegans_Training}).

\bigskip
\noindent\textbf{Motor control.} Both the model dynamics and the motor control were inspired by the GoPiGo project \citep{GoPiGo2024}, developed to leverage the \emph{C. elegans} connectome in order to control a Mindstorms robot.
Specifically, motion control is generated by summing the membrane potentials of the muscle cells that each motor neuron activates. Summing is calculated separately for the left and right sides. The accumulated activity for left and right motor movements are computed as follows:
\begin{align}
a_{\text{left}} = \sum_{m \in M_{\text{left}}} V_m,\qquad
a_{\text{right}} = \sum_{m \in M_{\text{right}}} V_m
\end{align}
where $M_{\text{left}}$ and $M_{\text{right}}$ are the sets of left and right muscle cells, respectively, and $V_m$ are the corresponding membrane potentials.
These accumulated inputs determine the speed of movement in the left and right directions.

\vskip2mm
\noindent\underline{\emph{Worm speed}}. 
The angular velocity of the worm is taken to be $\omega = k_a(a_{\text{left}} - a_{\text{right}})$,
where $k_a$ was set to 0.1 based on previous work \citep{GoPiGo2024}.
The linear velocity, in turn, is calculated as $v=k_l(a_{\text{left}} + a_{\text{right}})$, where $k_l$ was adjusted to be 
0.14, and the absolute velocity was limited between $v_\text{min}=10.7$ and $v_\text{max}=21.4$, for computational convenience and to qualitatively mimic the behavior of \emph{C. elegans} during a food foraging task \citep{GoPiGo2024}.

\vskip2mm
\noindent\underline{\emph{Direction update}}. The worm’s facing direction, given by the angle $\theta$, is updated by  the angular velocity, $\theta_{\text{new}} = (\theta_{\text{old}} + \omega)\bmod (2 \pi)$, where we take into account that the time step is $\Delta t=1$ (in arbitrary time units).

\vskip2mm
\noindent\underline{\emph{Position update}}. The new position $(x, y)$ of the worm is computed based on the updated linear velocity and facing direction:
\begin{equation}
x_{\text{new}} = x_{\text{old}} + v_{\text{adj}} \cdot \cos(\theta_{\text{new}}),\quad
y_{\text{new}} = y_{\text{old}} + v_{\text{adj}} \cdot \sin(\theta_{\text{new}})
\end{equation}
The position is constrained within the bounds of the simulation environment (see below for a description of the worm's environment).
To that end, if the distance between the worm and the wall is smaller than a fixed cutoff (here fixed to be 100 spatial units), a set of object avoidance sensory neurons (FLPR, FLPL, ASHL, ASHR, IL1VL, IL1VR, OLQDL, OLQDR, OLQVR, and OLQVL) is made to fire.
These specific neurons were selected based on the connectome implementation used in the GoPiGo project \citep{GoPiGo2024}.
In case the training procedure impairs the natural sensory avoidance system of the worm (through unsuitable changes of the corresponding synaptic weights), a failsafe condition prevents the worm from reaching values of $x$ and $y$ outside the environment, by fixing those coordinates to their boundary values when needed (absorbing boundary conditions).
This is rarely used, as the worm has collision avoidance circuitry is usually sufficient: food is not placed close to the wall, hence worms typically avoid the system boundaries.

\bigskip
\noindent\textbf{Environment}.
The simulation environment consists of a rectangular area of $1600 \times 1200$ spatial units, with the worms being initialized in the center of this space.
The worms navigate and interact with 36 distinct food sources, distributed in different patterns, for 250 time units. 
Food chemical sensing is a crucial way in which the worm interacts with its environment.
When the worm is closer to a food source than a specified detection range $D$ (here considered to be 150 spatial units), a set of food-related sensory neurons fire. The neurons involved in this response are ADFL, ADFR, ASGR, ASGL, ASIL, ASIR, ASJR, and ASJL.
These specific neurons were selected based on the connectome implementation used in the GoPiGo project \citep{GoPiGo2024}.
When the distance between the worm and the food source is smaller than a specified consumption range $C$ (here considered to be 20 spatial units), the work eats the food source (which thereby disappears) and receives a full reward (see Sec.~\ref{sec:fitness} below).

\subsection{The ENOMAD framework}
\label{sec:fitness}

As stated in Sec.~\ref{sec:intro} above, our approach combines a nonlinear optimization method, NOMAD, with evolutionary exploration.
In what follows we describe the different components of this hybrid framework.

\begin{figure}[htbp]
    \centering
    \includegraphics[width=0.8\textwidth]{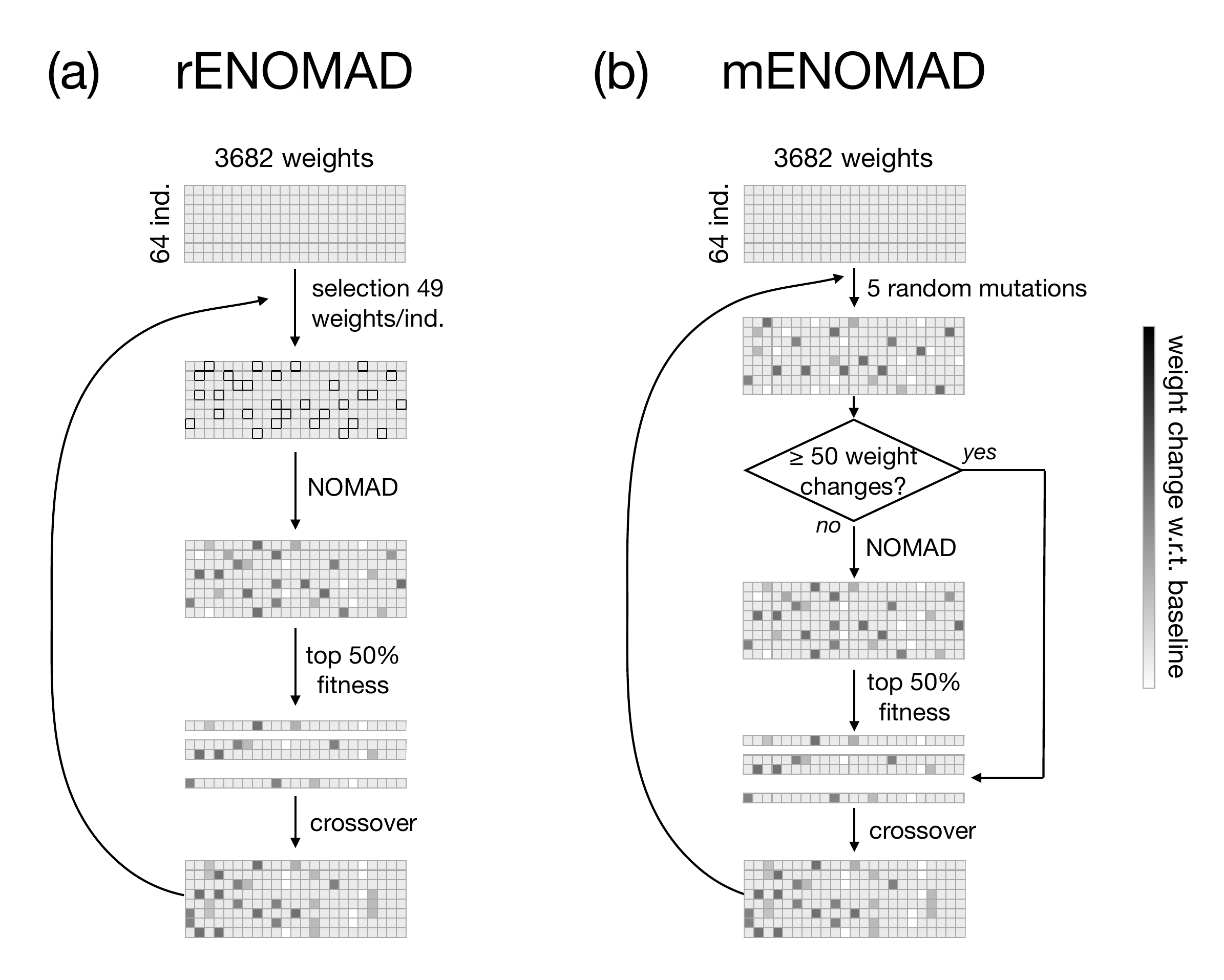}
    \caption{\textbf{Our two algorithmic variants of the ENOMAD pipeline}.
    The ``genome'' of the worm's neural network (the set of connection weights) is initialized to its biological prior, as shown in the top matrices in both panels.
    Light grey in those top matrices represents the basal biological values of the weights, as obtained from the c302 connectome \citep{c302}.
    After initialization, changes in the weights with respect to their biological levels are shown from white to black.
    (a) In rENOMAD, 49 weights are selected at random at each iteration, and NOMAD optimizes those weights, followed by fitness selection and crossover, after which a new iteration begins.
    (b) In mENOMAD, 5 mutations are added after each iteration.
    If the accumulated number of changes is 50 or larger, the NOMAD step is skipped, proceeding only with the evolutionary part of the pipeline.
    This sidestepping is rarely needed, as an our cardinality regularization heavily penalizes large numbers of weight changes. Hence the method wanders far away from the natural network.}
    \label{fig:scheme}
\end{figure}

\bigskip
\noindent\textbf{ENOMAD algorithmic variants}.
We have explored two distinct pipelines that combine the NOMAD algorithm with evolutionary approaches to optimize the \emph{C. elegans} connectome.
In the basal version of the method (Fig.~\ref{fig:scheme}a), which we call random ENOMAD (abbreviated as rENOMAD), 49 weights are selected at random for optimization by the NOMAD algorithm in each iteration.
This method relies on NOMAD’s ability to quickly solve high-dimensional black-box optimization problems.
We limit the number of optimized weights to preserve NOMAD’s convergence guarantee, which applies when optimizing less than 50 variables.
In turn, the mutation-based ENOMAD (which we call mENOMAD, see Fig.~\ref{fig:scheme}b) further limits the number of weight changes by mutating 5 weights per iteration prior to the optimization step.
In this version, the NOMAD routine is run only on the weights changed with respect to the original connectome, and only when the number of these weights is under 50, again for computational feasibility of the optimization method. 
rENOMAD allows more extensive exploration, leading to smaller sized changes in the weights.
However, this comes at the cost of changing the values of a higher number of weights, therefore moving farther away from the original connectome structure than mENOMAD.

\bigskip
\noindent\textbf{Population initialization}.
Our ``genome'' consists of an $S$-dimensional weight vector labeled $\mathbf{W}$, which determines the dynamics of our natural neural network and consequently the behavior of our worm (see Sec.~\ref{sec:model} above).
In our case $S=3682$ weights.
We initialize a population of $N=64$ worms with the same weight vector, corresponding to the biological weights coming from the c302 connectome dataset \citep{c302}.
This is shown in the top matrices of Fig.~\ref{fig:scheme}, where the homogeneous light grey shading represents no changes with respect to the biological prior.

\bigskip
\noindent\textbf{NOMAD optimization}.
NOMAD (Nonlinear Optimization by Mesh-Adaptive Direct search) is a derivative-free optimizer built on the MADS framework.
Each iteration evaluates trial points on a mesh around the current best solution.
If any trial improves the objective, the mesh is refined (step size shrinks) to exploit that region; otherwise it is coarsened (step size grows) to broaden the search.
Because it relies solely on function evaluations not gradients NOMAD copes well with noisy or black-box objectives and offers provable convergence to stationary points, provided each search is limited to less than 50 parameters.

\bigskip
\noindent\textbf{Fitness evaluation}.
To evaluate the worm's effectiveness at locating food sources within a simulation, we implemented a reward function designed to incentive certain realistic proximity-based behaviors.
The function computes the reward based on the worm's position relative to all food sources, incorporating two key parameters defined above (Sec.~\ref{sec:model}): detection range $D=150$ and consumption range $C=20$.
The reward $R(t)$ at time $t$ is calculated as follows:
\begin{equation}
R(t) = \sum_{\mathbf{f} \in \{\text{food}\}} \left\{
\begin{array}{ll}
R_\text{full} & \qquad\text{if } 0 < \| \mathbf{r}(t) - \mathbf{f} \| < C \\
R_\text{partial}\cdot \text{ReLU} \left(\frac{D - \| \mathbf{r}(t) - \mathbf{f} \|}{D}\right) 
& \qquad\text{if } C < \| \mathbf{r}(t) - \mathbf{f} \| < D
\end{array}
\right.
\end{equation}
Here $\{\text{food}\}$ is the set of food source positions,
$\mathbf{r}(t)$ is the worm's current position,
$R_\text{full}=30$ is the full reward given when the worm eats the food,
$R_\text{partial}=1/30$ is the maximum partial reward given when the worm detects the food, and
$\| \mathbf{r}(t) - \mathbf{f} \|$ represents the Euclidean distance between the worm and a food source.
The rectifier linear unit (ReLU) activation function is the identity if its argument is positive, and 0 if it is negative.

Once the rewards are calculated, we define fitness as the accumulated reward throughout the 250 time steps of the simulation:
\begin{equation}
f = \!\!\!\sum_{t \in \{250\}} \!\!\!R(t)
\end{equation}

\bigskip
\noindent\textbf{Cardinality regularization}.
To ensure that the evolved worm's genome (set of weights) remains close to the original connectome, we penalize genomes that deviate significantly from the original by subtracting the following regularizer (penalty) from the fitness \citep{louizos2018}:
\begin{equation}
  \mathcal{R}(\theta) = -\lambda \cdot |{\mathbf{W} \neq \mathbf{W}_0}|^{1.3},
  \label{eq:reg}
\end{equation}
where $\mathbf{W}$ is the candidate weight vector, $\mathbf{W}_0$ corresponds to the original connectome, and $|\cdots |$ represents the cardinality of the set of weights that are different from the original set.
The approach assumes that the original connectome has several strong local minima in its vicinity for the task proposed.
While there may be other local minima in the $\sim 4000$-dimensional search space, finding these can be time-consuming and is not guaranteed to yield strong local minima.
Our simulations show that nearby, highly performant local minima exist. 
Hence, our approach leverages the weights of the original connectome to inform our search, essentially using natural selection to direct our artificial evolutionary algorithm.
The supralinear exponent (1.3) in Eq.~\eqref{eq:reg} drives the search toward \textit{few} but meaningful edits.
In turn, the choice of \(\lambda\) influences the strength of the penalty: smaller \(\lambda\) relaxes the
constraint, while larger \(\lambda\) keeps solutions closer to the original network.
We set $\lambda = 0.1$; at 49 weight changes the penalty is equal to $\approx15$ (half the value of the consumption reward $C$).

\bigskip
\noindent\textbf{Crossover}.
We now turn to the evolutionary part of the framework, which is centered around the method of binary crossover.
Its primary purpose is to combine the genetic material of two parents to generate offspring that potentially inherit beneficial traits from both of them \citep{golberg1989}.
The process begins by removing the bottom 50th percentile of individuals. We then calculate the fitness probabilities of the top 50th percentile of individuals in the population ($p_i = f_i/\sum_{j} f_j$), and select two parents from the population according to these probabilities.
Finally, we compute the crossover probability based on the fitness of the selected parents, $p_c = p_{1}/(p_{1} + p_{2})$, which determines the weight matrix for the offspring per gene (weight) $i$:
   \begin{equation}
       W_{\text{offspring}{,i}} = 
       \begin{cases} 
       W_{1,i} & \text{with probability } p_c \\
       W_{2,i} & \text{with probability } 1 - p_c
       \end{cases}
   \end{equation}
   where $W_{1}$ and $W_{2}$ are the weight matrices of the two parents.

\bigskip
\noindent\textbf{Mutation}.
In contrast with the random version of ENOMAD (Fig.~\ref{fig:scheme}a), in which 49 weights are selected at random to be optimized using NOMAD in every iteration, the mutation-based version of ENOMAD (Fig.~\ref{fig:scheme}b) introduces constrained variability by replacing existing parameter values in the weight matrix (5 per iteration, in our case) with new values drawn from a uniform distribution.
The mutation function uses symmetric uniform bounds $\mathcal{U}(-20,20)$. Although the observed weights range from $-13.0$ to $37.0$, we cap the upper bound at 20 to limit the influence of rare large positive values (top 0.51\% of the distribution), which could otherwise dominate mutations and slow convergence. We also extend the lower bound slightly beyond the observed minimum to preserve symmetry and avoid sign bias. Using a uniform distribution promotes broad exploration, which is particularly advantageous during the initial stages of optimization.

\subsection{Baseline strategies}
\label{sec:baseline}

In the Results section, we compare the performance of our two variants of ENOMAD with three alternative approaches, which are described in what follows.

\bigskip
\noindent\textbf{OpenAI-ES}.
OpenAI Evolution Strategies (OpenAI-ES) is an alternative black-box optimization method: it samples parameter perturbations from a multivariate Gaussian, estimates the gradient of expected fitness with those samples, and applies a stochastic natural-gradient step \citep{Salimans2017OpenAIES}.
Because fitness evaluations are independent, the method scales almost linearly across hundreds or thousands of CPU cores, enabling fast, communication-light training that matches deep-RL performance on many benchmarks (potentially cite here).
This massively parallel design makes OpenAI-ES a practical choice for high-dimensional neuroevolution tasks.

\bigskip
\noindent\textbf{Evolutionary algorithm}.
We also compare ENOMAD with a pure evolutionary algorithm initialized at the original connectome, that does not use NOMAD for optimization.
The algorithm uses the cardinality regularizer described above, which penalizes deviations from the wild-type genome.
It also applies fitness-based uniform crossover for selection pressure, and mutates five random weights per offspring, as discussed above, to find improvements in fitness.

\bigskip
\noindent\textbf{Crossover-free NOMAD} (cfNOMAD).
As a minimalist control we pair \emph{two} uniformly drawn weight
mutations (applied once every four generations) with a NOMAD call optimizing the mutated genes. We keep the best performing genome each generation so that performance is not lost.
There is \emph{no} crossover, so improvements arise solely from mutations or local NOMAD
refinement. 
Because its mutation cadence matches the mENOMAD scheme yet lacks
selection pressure, Random-50 NOMAD provides a clean baseline for the
added value of our structured algorithms.

\subsection{Compute Environment}

All simulations were executed on a single workstation running Ubuntu~22.04 LTS equipped with an \textbf{Intel\textsuperscript{\textregistered} Core\textsuperscript{TM} i7-3770K CPU} (4 physical cores, 8 hardware threads, base 3.5 GHz) and \textbf{32 GB DDR3 RAM}.  
No GPU acceleration was used; every fitness evaluation ran on the CPU. Parallelism was obtained with \texttt{Ray~2.10} by launching one worker per logical core (8 workers), so a complete 30-seed sweep finished in \(\approx\) 20 minutes wall-clock time.  
The code was written in \texttt{Python 3.11.4}; key libraries: \texttt{NumPy 1.26}, \texttt{PyNomad 1.0.5}, \texttt{gymnasium 0.29}, and \texttt{matplotlib 3.9}.
An exact \texttt{conda} environment file is provided in the accompanying repository (\url{https://github.com/dsb-lab/C_Elegans_Training}).

\section{Results}

\begin{figure}[htbp]
  \centerline{\includegraphics[width=0.95\linewidth]{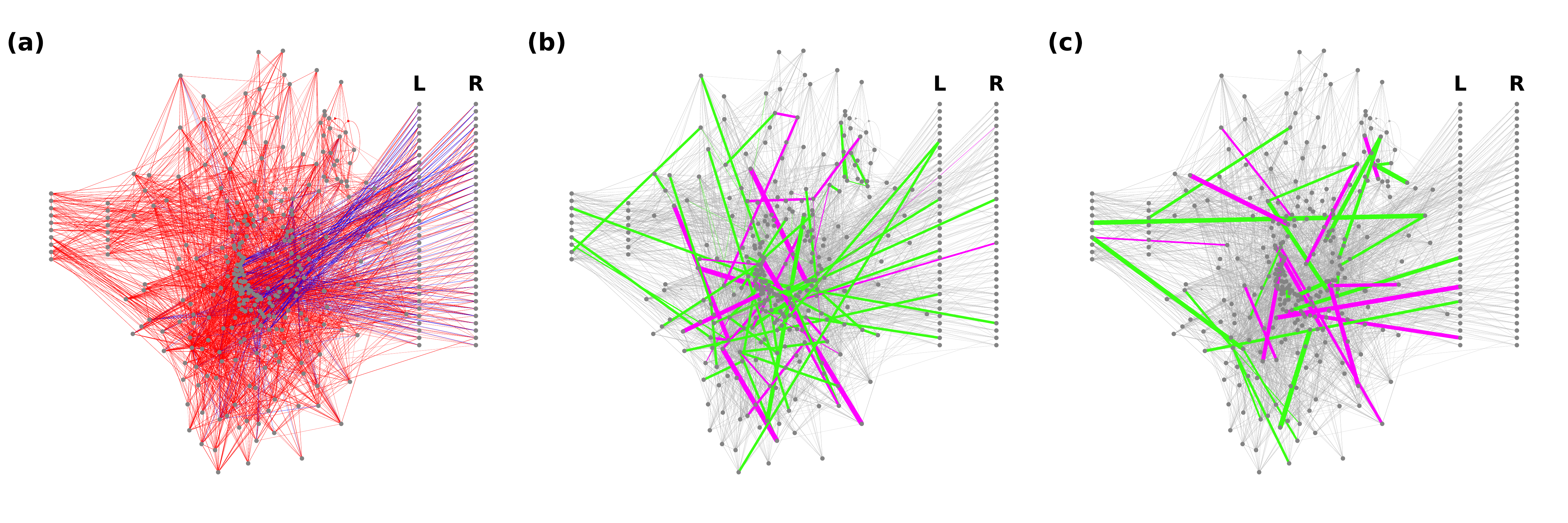}}
  \caption{
    \label{fig:networks}%
\textbf{Effect of ENOMAD on the biological connectome}. The original \emph{C. elegans} connectome is shown in panel (a), as obtained from the c302 project \citep{c302}.
The nodes are stratified according to their location in the networks, with purely upstream nodes placed at the left and purely downstream nodes placed at the right. Red lines denote excitatory synapses, blue lines denote inhibitory ones.
The width of the edges varies according to the values of the weights (synaptic connection strengths).
Panels (b) and (c) show the weights changed by the rENOMAD and mENOMAD methods, respectively, with weight decreases shown in green and weight increases shown in magenta, with the widths representing again the amount of change.
}
\end{figure}

We have introduced in Sec.~\ref{sec:framework} above a training strategy for highly recurrent biological networks, termed ENOMAD, which combines a nonlinear optimization method with an evolutionary algorithm.
We now evaluate the effectiveness of this learning framework on the \emph{C. elegans} connectome, using a suite of food-foraging benchmarks as a testbed.
Crucially, the method starts from existing biological knowledge coming from the c302 project \citep{c302}, which provides a biological prior in the form of a list of nodes (neurons and muscle cells innervated by neurons) and edges (connections between them), together with the weights of those edges (strengths of the synaptic connections).
Figure~\ref{fig:networks}(a) represents this biological prior, with the sensory neurons ($N_\text{in}=18$) forming two layers on the left, and the muscle cells ($N_\text{out}=68$) shown in two layers on the right. The muscle cells are fully downstream of the network and behave as actuators (34 control the left part of the worm's body, and 34 control the right part, shown and labeled separately in the plot). These actuator cells are controlled mainly by a large set of recurrently connected neurons ($N_\text{rec}=282$), which process the input information received by the upstream neurons by projecting it nonlinearly in a high-dimensional space \citep{buonomano2009,sussillo2009}.

We applied the two algorithmic variants of ENOMAD, described in Sec.~\ref{sec:fitness} above and represented schematically in Fig.~\ref{fig:scheme}, to the \emph{C. elegans} connectome, with the goal of optimizing the response of our virtual worm (Sec.~\ref{sec:model}) to a set of food-foraging tasks.
As discussed in Sec.~\ref{sec:fitness}, rENOMAD uses NOMAD on 49 random parameters for each member of the population every generation, followed by a crossover step to ensure the population converges towards optimal solutions found by individuals. This modifies many parameters but does not massively change the original values of the parameters; leading to strong performance with minimal changes in the values of the weights.
In turn, mENOMAD mutates a small subset ($< 50$) of weights, which are then fine-tuned by NOMAD, finishing with crossover to improve population convergence. This approach enables targeted adaptation while largely preserving the original connectome's parameters. 

\begin{figure}[htbp]
    \centering
\includegraphics[width=0.9\textwidth]{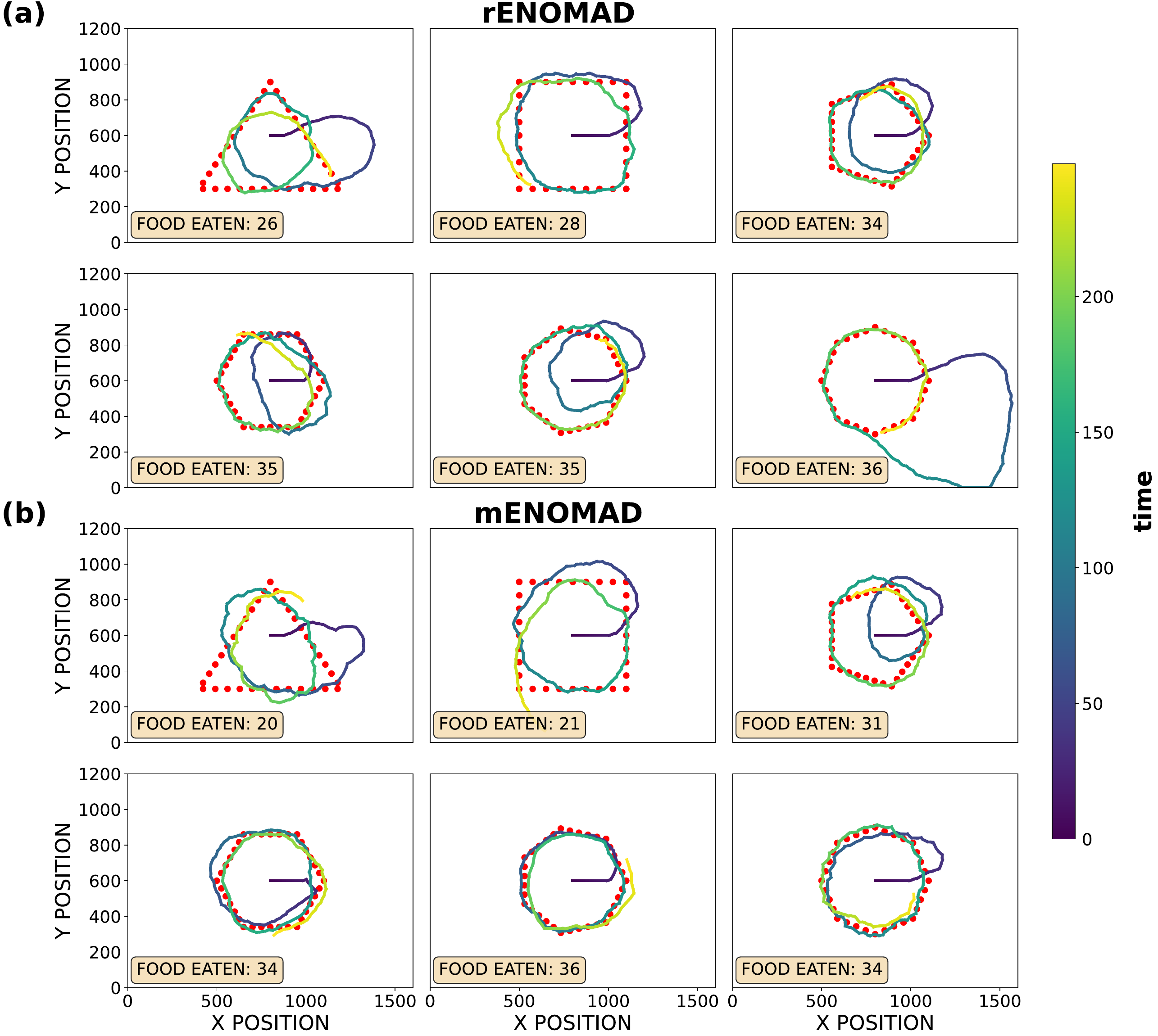}\\
    \caption{\textbf{Worm trajectories for a variety of food-foraging tasks}.
    The behavior of the worm with connectomes optimized by rENOMAD (a) and mENOMAD (b) is shown for six different tasks corresponding to food sources (red circles) placed in different polygonal patterns.
    The trajectories are colored as a function of time, with dark blue being the start of the simulation and yellow being the end.
    The initial food pattern is shown, but the food sources disappear after being eaten (see Supp. Videos 1 and 2).
    The training algorithm is run for 28 generations with rENOMAD and 280 with mENOMAD.}
    \label{fig:trajs}
\end{figure}

The effect of these two procedures on the set of synaptic weights of the neural network are shown in Fig.~\ref{fig:networks}(b,c).
As a result of the different ways in which the two methods sample the weights (see Sec.~\ref{sec:fitness}), the results confirm that rENOMAD changes a larger number of weights a small amount (Fig.~\ref{fig:networks}b), while mENOMAD maintains a larger fraction of the weights of the original connectome unchanged, but those that are changed are done so in larger amounts.
The corresponding behaviors of the worms for typical optimized connectomes arising from the two methods are shown in Fig.~\ref{fig:trajs} (see also Supp. Videos 1 and 2).
As can be seen in the plots, the worms quickly adapt to their environment, especially for polygons with higher numbers of sides, where less sharp turns are needed in order to track the food.

\begin{figure}[htbp]
\centerline{\includegraphics[width=0.95\linewidth]{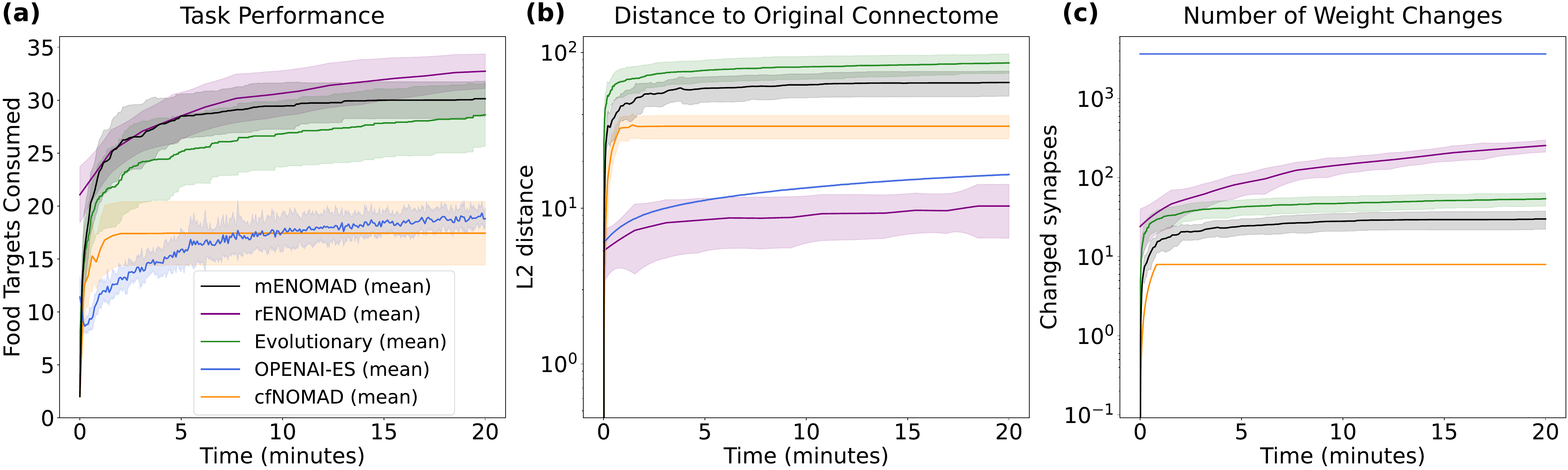}}
  \caption{%
    \label{fig:comp}%
\textbf{Benchmarking training algorithms for densely recurrent biological networks.}
The task performance (a), Euclidean ($L_2$) distance between the trained and original connectome (b), and weight-change counts ($L_0$ distance, c) are shown for five different reinforcement learning strategies as a function of computing time, for the pentagon food-gathering task.
Shadowing represents the standard deviation over 30 independent runs.
The training strategies compared here include our two algorithmic variants of ENOMAD, and the three baseline strategies described in Sec.~\ref{sec:baseline}.
Both rENOMAD and mENOMAD achieve the highest performance, with rENOMAD performing slightly better, but taking longer to reach maximum performance.
Both methods keep the closest distance--$L_2$ for rENOMAD and $L_0$ for mENOMAD--suggesting that the evolved weights are already near-optimal for the task. This indicates that our methods converge to highly effective local minima situated close to the original network.
Each algorithm incurs a different per-generation cost, hence the 20-minute window spans a different number of generations for each method (rENOMAD: 14 generations, mENOMAD: 190 generations, cfNOMAD: 100 generations, OpenAI-ES: 250 generations, pure evolutionary algorithm: 1,300 generations).
The training algorithm for rENOMAD and mENOMAD is run with half the computational budget given for ~\ref{fig:trajs}.
} \end{figure}

We next compared the performance of ENOMAD with other optimization methods, described in Sec.~\ref{sec:baseline} above.
To that end, we tracked the task performance (defined as the number of food sources eaten in a given time interval), the Euclidean distance from the original connectome (defined as $L_2=\| \mathbf{W} - \mathbf{W}_0 \|$), and the number of modified weights, as a function of the computing time. 
The results, show in Fig.~\ref{fig:comp}, allow us to conclude that both rENOMAD and mENOMAD surpass the performance of existing methods, while exhibiting the smallest Euclidean distance between trained and original connectome, and changing an intermediate number of weights, with respect to the the original connectome.
In particular, rENOMAD achieves the highest performance at the end of training, with 32.73 ± 1.66 food sources eaten.
In turn, mENOMAD also demonstrates strong performance and enables worms to consume, on average, over 13.6\% percent more food than worms trained using the evolutionary algorithm alone, while changing less than 1\% of the original connectome.
It is worth noticing that the results of Fig.~\ref{fig:comp} are shown as a function of computer clock time, not in terms of number of generations.

\begin{figure}[htbp]
    \centering
\includegraphics[width=0.95\textwidth]{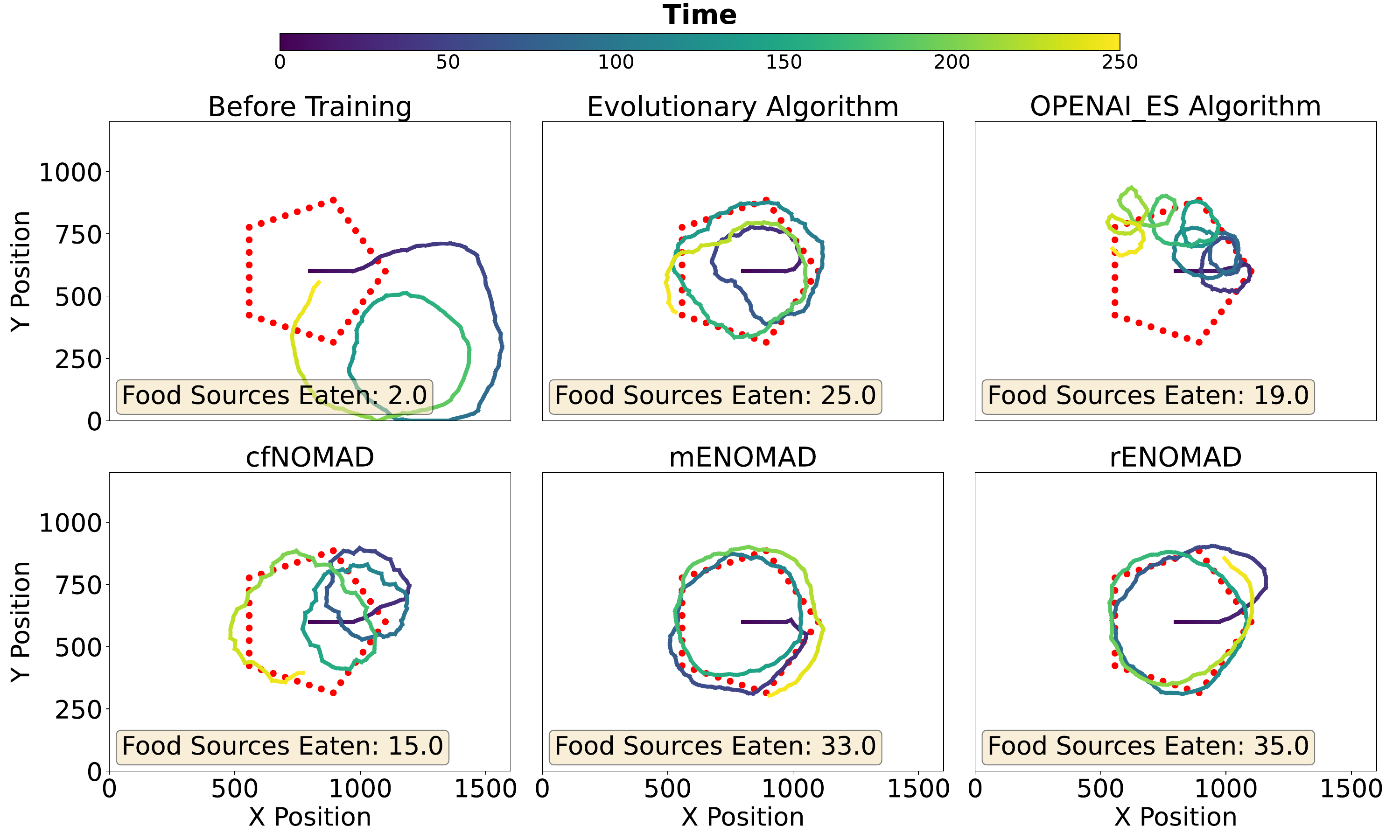}
\caption{\label{fig:comp_traj} \textbf{Comparing worm behavior after different training procedures.}
Worm trajectories are shown for the original connectome and five trained networks subject to a pentagon food-gathering task.
Red circles mark food targets and the trajectory's local color represents the time step of the simulation, with dark blue being the start and yellow the end.
The untrained connectome's path (top left) goes in smooth arcs that seldom intersect the pentagon, yielding low reward.
The evolutionary algorithm performs well but misses corners and lacks precision, a byproduct of using solely mutation to find fitness improvements. OpenAI-ES also improves the reward, but leads to small unproductive loops.
cfNOMAD's search also results in unproductive loops, leading to extended periods away from food.
mENOMAD and rENOMAD both reduce off-target excursions and enable the capturing of the majority of food targets.}
\end{figure}

The resulting behavior of the worms for the pentagonal task after training by all methods is shown in Fig.~\ref{fig:comp_traj}.
Consistent with the benchmarking results shown in Fig.~\ref{fig:comp}, the worm trajectories from ENOMAD-trained connectomes converge toward food source locations.
It is worth contrasting the behaviors of those optimal solutions with that of the untrained connectome (top left panel).
The results clearly show that the evolved connectome is not adapted specifically for the proposed task. In the opposite limit--in which a local minimum is found far from the original connectome--the resulting fitness after training is usually significantly lower compared to training near the original connectome (see Suppl. Fig.~1).

Take together, these results show that minor modifications to the biological weights, as uncovered by ENOMAD, lead to significantly improved task performance.
By initializing from the generalized original connectome and subsequently adapting the weights to a specific task, our method constitutes an instance of transfer learning within a biological network.

\section{Discussion}

Our results show that coupling an evolutionary outer loop with a NOMAD-based inner optimizer offers an efficient way to specialize densely recurrent natural circuits.
We have introduced two algorithmic variants of the approach, both of which--rENOMAD, which refines 49 randomly chosen synapses each generation, and mENOMAD, which first mutates $< 50$ targeted weights before refinement--outperformed other machine learning algorithms, such as OpenAI-ES and a lasso-regularized evolutionary baseline, on every food-foraging task tested with the \emph{C. elegans} connectome (Figs.~\ref{fig:comp} and \ref{fig:comp_traj}).

Our method remains both scalable and broadly transferable due to two key design choices:
\begin{enumerate}
    \item \textbf{Architecture-agnostic fitness calls.}  
          The optimizer treats the network and its environment as a black box, hence no gradients or Jacobians are required.
          Replacing the \emph{C. elegans} connectome with a fly visual circuit--or any mapped recurrent network--only demands a new fitness function.
\item \textbf{Fitness evaluation dominates runtime.}
Profiling shows that rENOMAD spends 99.78\% of its wall-clock time on worker processes. Of this, 71.83\% is consumed by the environment and neuronal simulations that compute fitness, 14.70\% by the \texttt{PyNomad} local search, and 13.25\% by parallel-overhead such as serialization, task scheduling, and workers idling for the slowest individual in a generation. Core genetic bookkeeping--selection, crossover, and logging--takes less than 0.2\%. 
Because the heavy workload (worker processes) is highly parallel, total wall-clock time decreases almost linearly as the CPU core count increases.
\end{enumerate}

Conventional machine-learning pipelines start with minimal prior structure and train a model to solve many tasks from scratch.
Our framework moves in the opposite direction: we begin with a well structured, pre-generalized biologically derived network and specialize it for a subset of behaviors.
Our results can thus be interpreted as an instantiation of transfer learning.
This approach excels when two conditions are met:
\begin{itemize}
    \item \textbf{Behavioral alignment} -- the target task is similar to actions the organism already performs.
    We can make this assumption in our case because the food collection problems we train the worm on are similar to tasks the worm performs in its natural habitat \citep{Chalasani2007}.
    \item \textbf{Strong parameter priors} -- the connectome's weights lie close to a good local optimum, courtesy of evolution.
    This is a crucial assumption as it allows us to use the NOMAD algorithm.
    If the original connectome is poorly optimized for the problem, transfer learning becomes much more challenging.
    This assumption also allows us to add cardinality regularization, where we penalize the worm having a weight vector with large differences from the original connectome.
\end{itemize}
Under these premises the optimizer can achieve large performance gains through small, local weight adjustments. When the task departs too far from the animal’s natural repertoire, the initial weights are unlikely to be near any high-quality minimum; the number of required changes grows, and NOMAD-based refinements lose their advantage over full retraining.

In summary, our work reveals that combining nonlinear optimization with evolutionary search is a powerful tool for training highly recurrent natural networks, when strong priors exist.
To build upon our study, future work should focus on refining the weight selection phase, potentially by incorporating techniques from network theory to identify the connections most critical to optimize.
In any case, applying ENOMAD to natural networks should allow researchers to fine tune approximate parameters of highly recurrent networks for a specific task when good priors are available.
In that way, these results could hopefully both push the boundaries of biological network research and harness natural evolutionary processes to guide the advancement of artificial intelligence.


%
%
%
%
%
%

\section*{Code availability}
All original code has been deposited at Github at the following repository: \url{https://github.com/dsb-lab/CElegans_Training}.

\section*{Acknowledgments}

This work was supported by the Ministerio de Ciencia, Innovaci\'on y Universidades and the Agencia Estatal de Innovación / FEDER (Spain) under project PID2024-160263NB-I00, and by the European Research Council, under Synergy grant 101167121 (CeLEARN).
M.W.C. was supported by a Summer Internship Award from the Center for Life Beyond Reed (CLBR).
We thank Anne Churchland, Gabriel Torregrosa, Raul de Palma, and the members of UPF's Dynamical Systems Biology lab for crucial insights and feedback throughout the project.
We also thank the Barcelona Program for Interdisciplinary Studies (BaPIS) at Pompeu Fabra University for bringing us together to work on this project.
Lastly, we wish to acknowledge the scientific, mathematical, and computational communities whose advancements have made this project possible.

%

\section*{Declaration of interests}

The authors declare no competing interests.

\section*{Supplementary information index}

\begin{description}
  \item Figure S1. Benchmarking including a variant of rENOMAD (shown in red) in which weights are initialized randomly--rather than from the biological connectome--using a uniform distribution over $(-20,20)$. This highlights that searches beginning from the original connectome outperform those starting far from it. \item Movie S1. Worm behavior after rENOMAD training for the pentagon food task. The activity of the network is shown on the right, with orange circles indicating neuron firing.
  \item Movie S2. Worm behavior after mENOMAD training for the pentagon food task. The activity of the network is shown on the right, with orange circles indicating neuron firing.
  \item Movie S3. Worm behavior of the untrained connectome for the pentagon food task. The activity of the network is shown on the right, with orange circles indicating neuron firing.
\end{description}

\bibliography{nomad}

\end{document}